\documentclass[10pt,twocolumn]{article}

\usepackage{times}
\usepackage{epsfig}
\usepackage{graphicx}
\usepackage{amsmath}
\usepackage{amssymb}
\usepackage{booktabs} 
\usepackage{multirow}
\usepackage{multicol}
\usepackage{array}
\usepackage{longtable}
\usepackage{comment}
\usepackage{ifthen}
\newboolean{markup_red}
\setboolean{markup_red}{true}   
\ifthenelse{\boolean{markup_red}}
{}
{}
\newboolean{markup_blue}
\setboolean{markup_blue}{false}   
\ifthenelse{\boolean{markup_blue}}
{}
{}

\usepackage[pagebackref=true,breaklinks=true,letterpaper=true,colorlinks,bookmarks=false]{hyperref}

\begin{document}

\title{DichroGAN: Towards Restoration of in-air Colours of Seafloor from Satellite Imagery}

\author{Salma  Gonz{\'a}lez-Sabbagh\footnote{Corresponding author}\\
\small Deakin University\\
\small Waurn Ponds, VIC, 3216, Australia\\
{\tt\small s.gonzalezsabbagh@research.deakin.edu.au}
\and
Antonio Robles-Kelly\\
\small The University of Adelaide\\
\small Adelaide, SA, 5005, Australia\\
{\tt\small antonio.robles-kelly@adelaide.edu.au }
\and
Shang Gao\\
\small Deakin University\\
\small Waurn Ponds, VIC, 3216, Australia\\
{\tt\small shang.gao@deakin.edu.au}
}
\date{}
\maketitle

\begin{abstract}
Recovering the in-air colours of seafloor from satellite imagery is a challenging task due to the exponential attenuation of light with depth in the water column. In this study, we present DichroGAN, a conditional generative adversarial network (cGAN) designed for this purpose. DichroGAN employs a two-steps simultaneous training: first, two generators utilise a hyperspectral image cube to estimate diffuse and specular reflections, thereby obtaining atmospheric scene radiance. Next, a third generator receives as input the generated scene radiance containing the features of each spectral band, while a fourth generator estimates the underwater light transmission. These generators work together to remove the effects of light absorption and scattering, restoring the in-air colours of seafloor based on the underwater image formation equation.
DichroGAN is trained on a compact dataset derived from PRISMA satellite imagery, comprising RGB images paired with their corresponding spectral bands and masks. Extensive experiments on both satellite and underwater datasets demonstrate that DichroGAN achieves competitive performance compared to state-of-the-art underwater restoration techniques.

\end{abstract}

\section{Introduction}
\label{sec:intro}
Underwater exploration currently presents two major challenges. First, the inherent risks associated with submersion make direct exploration and data collection difficult, resulting in limited and small-size datasets \cite{wang2023underwater,cui2024application}. Second, the optical properties of the water column cause light attenuation due to absorption and scattering, degrading most underwater imagery and inducing colour casts \cite{chang2023uidef}.
Thanks to advances in sensor technology and imaging platforms, underwater imagery is now more accessible \cite{marouchos2018challenges}. Extensive research has been conducted to improve underwater image quality through colour correction and by increasing datasets size.  This is not surprising since it has been shown that combining colour correction techniques with high-level vision tasks such as classification \cite{boone2022marine} can improve overall performance, when over-enhancement is avoided \cite{li2019underwater,wangandli2023underwater}. Many methods have been proposed for underwater image enhancement \cite{cong2023pugan,mishra2024u}, restoration \cite{Akkaynak_2019_CVPR,li2024cascaded}, depth estimation \cite{drews2016underwater,gupta2019unsupervised}, and image synthesis \cite{ueda2019underwater,desai2024rsuigm}. However, these approaches often struggle to generalise across different water types while removing colour distortions and recovering true in-air colours of submerged objects and seafloor features across large areas.

One approach to recovering the reflectance of underwater surfaces is remote sensing (RS). As a key application of RS, satellite imagery provides essential input for recovering the in-air colours of seafloor and facilitates large-scale environmental monitoring without the need of submersion. 

Note that satellite and underwater imagery follow different processing techniques. Satellite images deal with the effects of atmospheric column (e.g., ozone, water vapour) and water column (e.g., water molecules, suspended particles), which are typically corrected through atmospheric and water column correction techniques \cite{ong2021potential}. In contrast, underwater images primarily address light attenuation and scattering within the water column, which leads to colour distortions. Specifically, light attenuation in satellite imagery is often estimated using linear regression between seafloor reflectance and water column depth \cite{lyzenga1978passive}, while underwater image formation assumes an exponential decay of light intensity with distance \cite{Duntley:1963}.

\begin{figure}[t]
  \centering
   \includegraphics[width=1\linewidth]{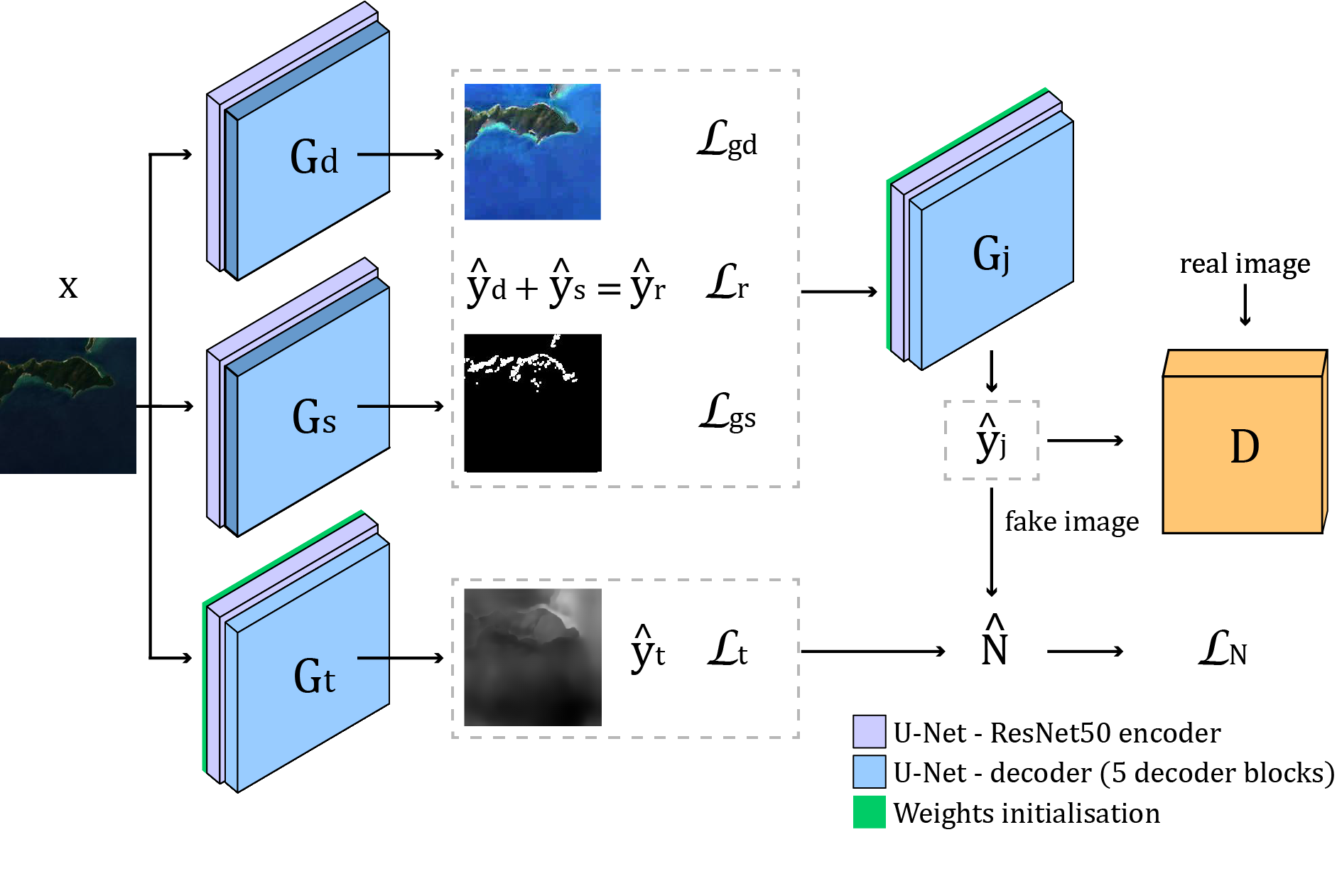}
   \caption{Overview of DichroGAN. It comprises 4 generators and 1 discriminator. Satellite scene radiance is obtained by summing diffuse and specular reflections generated by $G_{d}$ and $G_{s}$, respectively. Generated radiance serves as input for $G_{j}$, while $G_{t}$ generates depth map to estimate light transmission. $G_{j}$ and $G_{t}$ compute  UIFM \cite{Duntley:1963} to remove water column and recover in-air colours. Discriminator $D$ classifies between generated and real images.}
   \label{fig:Overview of our method}
\vspace{-0.03cm}
\end{figure}

In this work, we integrate diffuse and specular reflections into an underwater image formation model (UIFM) \cite{Duntley:1963} to remove the inherent optical effects of the water column. To achieve this, we introduce DichroGAN, a conditional generative adversarial network (cGAN) that first derives satellite scene radiance from these reflections, then corrects for water column effects and restores the in-air colours of the seafloor. Fig. \ref{fig:Overview of our method} illustrates its architecture.

Our main contributions are as follows: 

(1) We propose DichroGAN, the first cGAN designed to remove the water column and recover in-air seafloor colours from satellite imagery \footnote{The code and dataset are available at: \url{https://github.com/SalPGS/DichroGAN}}. Our framework employs four generators: two leverage diffuse and specular reflections from an hyperspectral image cube to solve the dichromatic reflection model, recovering seafloor albedo and downwelling light spectrum, while the other two estimate seafloor radiance and light transmission by solving the UIFM through transfer learning.

(2) We validate our method through extensive experiments and demonstrate competitive performance compared with state-of-the-art (SOTA) underwater techniques.

\section{Related Work}
\label{sec:related work}
Restoring the in-air colours of underwater scenes typically follow either an empirical or a physics-based approach using traditional or deep learning algorithms. 

\textbf{Empirical approaches}. Fu et al. \cite{fu2014retinex} enhanced underwater images by using colour correction to remove colour casts while applying Retinex theory to decompose reflectance and illumination. Luo et al. \cite{luo2021underwater} proposed an underwater image restoration and dehazing algorithm based on colour balance, contrast optimisation, and histogram stretching. Their method maintained the RGB channels intensity distribution to alleviate red light absorption. Similarly, Iqbal et al. \cite{iqbal2010enhancing} equalised  pixel values in the RGB and HSV colour spaces to remove colour casts while improving contrast and illumination. 

\textbf{Traditional algorithms.} Carlevaris-Bianco et al. \cite{carlevaris2010initial} proposed a prior for depth estimation and underwater dehazing by analysing RGB channel intensities. Based on the Dark Channel Prior (DCP) \cite{he2010single}, Drews et al. \cite{drews2016underwater} developed the Underwater DCP (UDCP) to estimate depth and restore underwater images by considering only the green and blue channel intensities. Peng et al. \cite{peng2017underwater} estimated depth using blurriness and light absorption to restore underwater radiance. The Sea-thru method  \cite{Akkaynak_2019_CVPR} removed water from underwater scenes based on a revised UIFM \cite{akkaynak2018revised} that incorporates the diffuse downwelling attenuation coefficient.

\textbf{Deep learning algorithms.} 
Li et al. \cite{li2019underwater} presented Water-Net, a convolutional neural network (CNN) that enhances underwater images using a fusion approach involving white balance, histogram equalisation, and gamma correction. Han et al. \cite{han2022underwater} developed Contrastive Underwater Restoration (CWR), a cGAN method inspired by \cite{isola2017image}, featuring a ResNet encoder-decoder generator and a two-layer perceptron generator, with a PatchGAN discriminator. Guo et al. \cite{guo2023underwater} presented URanker, a Transformer-based ranker for assessing underwater image quality. They then trained NU$^2$Net, a u-shaped network, using a pretrained URanker loss for underwater image enhancement. More recently, Khan et al. \cite{Khan_2025_WACV} extracted  phase image features to restore underwater images using a Transformer-based network named Phaseformer.

Most of the aforementioned methods rely on underwater RGB imagery. Some studies use satellite imagery, but they typically focus on in-situ or satellite spectral measurements to estimate absorption and scattering properties in the water column and seafloor reflectance \cite{lyzenga1978passive,mueller2003ocean,reichstetter2015bottom,mckinna2015semianalytical}. Other research efforts focus on haze removal \cite{kulkarni2023aerial}, cloud detection \cite{fraser2009method}, or water region extraction \cite{zhang2018automatic} from satellite imagery. However, research on recovering the in-air RGB colours of seafloor from satellite imagery remains fairly unexplored.

\section{Method}
\subsection{Light propagation in atmosphere}

A light source $L(\lambda_{i})$ illuminates an object with surface radiance $I(u, \lambda_{i})$, which reflects a spectrum $S(u, \lambda_{i})$ that encodes the physical and chemical properties of the object. Based on Shafer's dichromatic reflection model \cite{shafer:1985using}, the radiance emitted by the object can be expressed as a linear combination of its diffuse and specular reflections across the electromagnetic spectrum, incorporating indexed $i$ wavelengths $\lambda$, i.e., $\lambda_i \in \{\lambda_1, \ldots, \lambda_m\}$, where $m$ is the number of spectral bands:
\begin{equation}
  I(u, \lambda_{i}) = g(u)L(\lambda_{i})S(u, \lambda_{i})+k(u)L(\lambda_{i}),
  \label{eq:dichromatic reflectance equation}
\end{equation}
where $u$ is the pixel location, $g(u)$ represents the shading coefficient, and $k(u)$ represents the specular coefficient. Both coefficients depend on the light direction, viewpoint geometry, and surface position. 

\subsection{Light propagation in water}
Light propagation in water bodies is mostly affected by attenuation $\alpha$, which is wavelength dependent and defined as the sum of absorption $a$ and scattering $b$:

\begin{equation}
\alpha(\lambda_{i}) = a_{pdm}(\lambda_{i}) + b_{pdm}(\lambda_{i}),
\end{equation}
where $pdm$ describes the concentrations of particulate and dissolved matter, such as water molecules, salts, and chlorophyll. Attenuation also varies with water depth $z$, the distance from the object to the observer $r$, and viewing angles (zenith $\theta$, azimuth $\phi$) \cite{Duntley:1963,Mobley:1994}.

Since light propagation is medium-dependent, we adopt the UIFM from Duntley \cite{Duntley:1963} to estimate the underwater object reflection $N(z,\theta,\phi,\lambda_{i})$, given by:

\begin{equation}
\begin{split}
    {N}(z,\theta,\phi,\lambda_{i})={J}(z,\theta,\phi,\lambda_{i})\exp[-\alpha(z,\lambda_{i})r&]
    \\ + V(z,\theta,\phi,\lambda_{i})\exp[K(z,\theta,\phi,\lambda_{i})r \cos{(\theta)}]\\
    \times\{1-\exp[-\alpha(z,\lambda_{i})r+K(z,\theta,\phi,\lambda_{i})r \cos{(\theta)}&]\},
\end{split}
\label{eq: underwater image formation equation}
\end{equation}  
where $J(z,\theta,\phi,\lambda_{i}$) represents the object radiance, $V(z,\theta,\phi,\lambda_{i})$ represents the veiling light, and $K(z,\theta,\phi,\lambda_{i})$ is the diffuse attenuation coefficient. 

To simplify Eq. \ref{eq: underwater image formation equation}, we reformulate it in the image space using element-wise matrix operations, we have:

\begin{equation}
\begin{split}
    \mathbf{N}(u, \lambda_{i})=\mathbf{J}(u, \lambda_{i})\mathbf{T}(u, \lambda_{i})\\ +\mathbf{V}(u, \lambda_{i}) (1- \mathbf{T}(u, \lambda_{i})),
\label{eq: simplifyed underwater image formation equation}
\end{split}
\end{equation}  
where light transmission $\mathbf{T}(u, \lambda_{i})$ describes the exponential attenuation of light through the water column:
\begin{equation}
\mathbf{T}(u, \lambda_{i}) = \exp[-r\alpha(z,\lambda_{i})].
\end{equation} 

Assuming an overhead (nadir) viewpoint with a zenith angle $\cos(\theta) = 0$, the diffuse attenuation coefficient $K(z,\theta,\phi,\lambda_{i})$ approaches zero. This simplifies Eq. \ref{eq: simplifyed underwater image formation equation} and allows us to remove water column optical properties using: 

\begin{equation}
    \mathbf{J}(u, \lambda_{i}) = \left[ \frac{\mathbf{N}(u, \lambda_{i}) - \mathbf{V}(u, \lambda_{i})}{\mathbf{T}(u, \lambda_{i})} \right] + \mathbf{V}(u, \lambda_{i}).
    \label{eq: solving for R}
\end{equation}

Note that by removing the water column,  $\mathbf{J}(u, \lambda_{i})$ becomes $\mathbf{I}(u, \lambda_{i})$ from Eq.~\ref{eq:dichromatic reflectance equation} which includes the seafloor albedo and the downwelling light spectrum. Our method seeks to explicitly separate the diffuse and specular reflections by integrating the dichromatic reflection model.

\subsection{DichroGAN network}

We follow an unsupervised learning approach to address the inherent ambiguities of the aquatic medium by first separating observed radiance into its fundamental components: the illuminant spectrum and object reflectances. We then correct for water column distortions caused by absorption and scattering while preserving reflectance physics to recover surface albedo and the in-air radiance of seafloor.

In our proposed DichroGAN, we extend the standard cGAN \cite{mirza2014conditional,isola2017image} by employing four generators and a single discriminator $D$. Generators $G_{d}$, $G_{s}$, and $G_{t}$ take an RGB satellite image as input $x$, while $G_{j}$ receives as input $\hat{y}_r$, an RGB prediction generated by $G_{d} + G_{s}$, thereby learning the mapping $(\hat{y}_r, z) \rightarrow{\hat{y}}$. The adversarial objective function for the cGAN is given by:

\begin{equation}
\begin{split}
\label{eq: cGAN loss}
\min_{\theta_{g}} \max_{\theta_{d}} \mathcal{L}_{cGAN}=\mathbb{E}_{x,y}\left[\log D(x, y;\theta_{d})\right] + \\\mathbb{E}_{x, z} \left[\log(1- D(x,\theta_{d}, G_{j}(\hat{y}_r,z;\theta_{g}))\right],\\
\end{split}
\end{equation}
where $y$ is the recovered radiance, as defined in Eq.~\ref{eq: solving for R}, and $\theta_{g}$ and $\theta_{d}$ are the generator and discriminator parameters, respectively.

\textbf{Light source estimation.} Our first objective is to recover the seafloor radiance in water-covered regions. We begin by computing the surface radiance using Eq.~\ref{eq:dichromatic reflectance equation}, estimating diffuse and specular reflections via generators $G_{d}$ and $G_{s}$, respectively. To address illumination, we  focus on the light source, and assume the Grey World (GW) \cite{ebner:2007} hypothesis. We estimate a homogeneous illuminant spectrum $L(\lambda_{i})$ from the average scene colour in a  hyperspectral image cube $\textit{Im}$ containing $63$ indexed spectral bands $\lambda_{i}$:

\begin{equation}
    L_{gw}(\lambda_{i})= \frac{1}{n} \sum_{u\in \textit{Im}}I(u),
    \label{eq: illuminant grewy world}
\end{equation}
where $n$ is the total number of pixels. 
From Eq. \ref{eq:dichromatic reflectance equation} we can compute $k(u)$ as $\frac{I(u, \lambda_i)}{L(\lambda_i)}$. Thus, the expected value of $k(u)$ across the wavelengths can be obtained based on the illumination spectrum and medium properties:

\begin{equation}
  \mathbb{E}[k(u)] = \frac{1}{m} \sum_{i=1}^{m} \frac{I(u, \lambda_i)}{L_{gw}(\lambda_i)}. \quad 
\end{equation}

We then solve for the shading factor and surface reflectance where $gS(u,\lambda_{i}) = g(u)S(u,\lambda_{i})$:

\begin{equation}
    gS(u,\lambda_{i}) = \frac{I(u, \lambda_{i})}{L_{gw}(\lambda_{i})} - \mathbb{E}[k(u)].
     \label{eq: solve for shading and surface reflectance}
\end{equation}

\textbf{Separation of diffuse and specular reflections.} Using $G_{d}$ and $G_{s}$, we estimate diffuse and specular reflections. First, we implement a linear histogram stretch to extract the RGB values of the diffuse and specular reflections. Next, we minimise the loss for $G_{d}$ where $\hat{y}_{d} = G_{d}(x, z; \theta_{gd})$. The loss function for the diffuse reflectance is formulated as: 

\begin{equation}
    \mathcal{L}_{gd} = \|L_{gw}(\lambda)g(u)S(u,\lambda)-\hat{y}_{d} \|_{1}.
    \label{eq: loss diffuse reflectance}
\end{equation}

Similarly, we estimate specular reflectance through $G_{s}$, where $\hat{y}_{s} = G_{s}(x, z; \theta_{gs})$. The loss function for the specular reflectance is:

\begin{equation}
    \mathcal{L}_{gs} = \|k(u)L_{gw}(\lambda)-\hat{y}_{s}\|_{1}.
    \label{eq: loss specular reflection}
\end{equation}

With these elements at hand, we are able to recover the radiance by minimising the error between the hyperspectral image cube and  the dichromatic model reconstruction derived from $G_{d}$ and $G_{s}$. Using the $L2$ norm  \cite{huynh2010solution}, we define: 

\begin{equation}
    \mathcal{L}_{r} = 
    m \|I(u,\lambda)- G_{s} + G_{d}\|_{2}.
    \label{eq: loss dichromatic model}
\end{equation}

\begin{figure}
        \centering
        \begin{tabular}{ccccc}
        \multirow{8}{1em}{\rotatebox[origin=l]{90}{\scriptsize PRISMA}} &
            \includegraphics[width=6.8em]{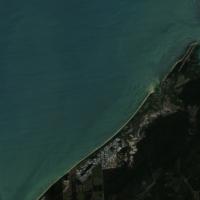}& 
            \includegraphics[width=6.8em]{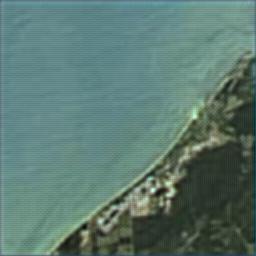}& 
            \includegraphics[width=6.8em]{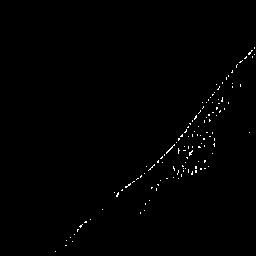}\\ 
            &
            \includegraphics[width=6.8em]{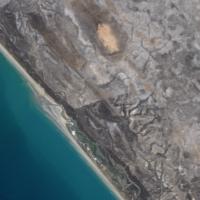}& 
            \includegraphics[width=6.8em]{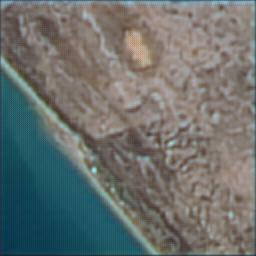} & 
            \includegraphics[width=6.8em]{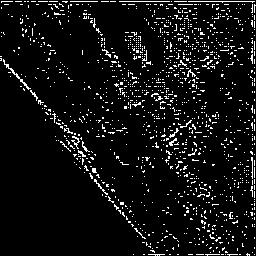} \\
            &
            \includegraphics[width=6.8em]{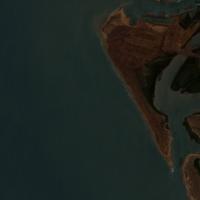}& 
            \includegraphics[width=6.8em]{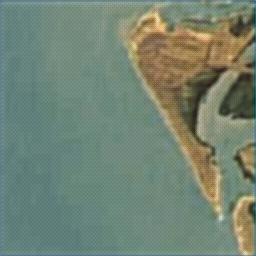}&
            \includegraphics[width=6.8em]{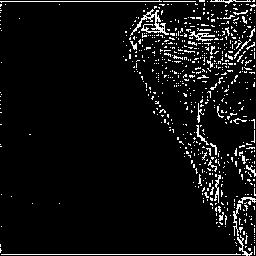}\\

             &\scriptsize Input image &\scriptsize Diffuse  & \scriptsize Specular \\
             &\scriptsize  &\scriptsize reflectance & \scriptsize reflectance 
        
        \end{tabular}
        \vspace{0.05cm}
        \caption{Sample results of generated diffuse and specular reflections on PRISMA test dataset.}
	\label{fig: diffuse and specular examples}
    \vspace{-0.3cm}
    \end{figure}

Note that our interest lies in the features of regions covered by water, thus we ignore land regions by applying a mask $m$ which penalises only the predictions within the water-covered areas. Examples of generated diffuse and specular reflections are shown in Fig.~\ref{fig: diffuse and specular examples}.

\textbf{Veiling and depth map estimation.} We then focus on the water medium by incorporate the recovered radiance into the UIFM (Eq. \ref{eq: simplifyed underwater image formation equation}). The veiling light $V_{gw}(u,\lambda_{i})$ is obtained via GW, while light transmission $T(u,\lambda_{i})$ is estimated using the pretrained method in \cite{gonzalez2025scene}. 

The attenuation coefficient is set to $0.9$, assuming deep-water and complete wavelength dependency. The loss function for $G_{t}$, denoted as $t$, is:

\begin{equation}
    t = m\|T(u,\lambda) - \hat{y}_t)\|_{1},
    \label{eq: loss transmission}
\end{equation}
where $\hat{y}_{t}$ is predicted by $G_{t}(x, z; \theta_{gt})$. Since our focus is on features of water-covered regions, land areas are masked using $m$.
To further constrain depth estimation, we adopt the scale-invariant loss from \cite{eigen2015predicting}: 

\begin{equation}
\mathcal{L}_{t}= \log(t+0.5) + \frac{1}{n}\sum_{i=1}(\nabla_{x}t_{i} + \nabla_{\hat{y}_{t}}t_{i}).
\label{eq: depth loss}
\end{equation}

Generator $G_{j}$ receives the predicted radiance $\hat{y}_r = G_{s} + G_{d}$ and removes water column distortions by minimising:

 \begin{equation}
    \mathcal{L}_{gj} = \sum^{n}_{i=1} m \|\mathbf{R}(u,\lambda)- \hat{y}_{j})\|_{1},
    \label{eq: loss generator raaciance}
\end{equation}
where $\hat{y}_{j}$ is predicted by $G_{j}(\hat{y}_r, z; \theta_{gj})$. 

To further constrain the training, we synthesise a fake underwater image by applying Eq.~\ref{eq: simplifyed underwater image formation equation} in the generated domain: 
\begin{equation}
 \mathbf{\hat{N}}(u, \lambda)= \mathbf{\hat{y}_{j}}\mathbf{\hat{y}_{t}} + \mathbf{V_{gw}}(u,\lambda)(1-\mathbf{\hat{y}_{t}}). 
\end{equation}

The corresponding loss function for the generated underwater image is:

 \begin{equation}
    \mathcal{L}_{N} = m \|\mathbf{N}(u,\lambda)- \mathbf{\hat{N}}(u, \lambda))\|_{1}.
    \label{eq: loss generator raaciance}
\end{equation}

Finally, the overall loss for DichroGan is:
 \begin{equation}
 \begin{split}
    \mathcal{L}_{obj} = \min_{\theta_{g}}\max_{\theta_{d}} \mathcal{L}_{cGAN} + \gamma(\mathcal{L}_{gs}&  + \mathcal{L}_{gd}) \\ + \sigma\mathcal{L}_{r} + \iota\mathcal{L}_{gj} + \tau\mathcal{L}_{t} + \nu\mathcal{L}_{N},
    \label{eq: objective loss}
    \end{split}
\end{equation}
where $\gamma, \sigma, \iota $, $\tau$, and $\nu$ are weights to balance the loss terms.


\setlength{\tabcolsep}{0.99pt}
\renewcommand{\arraystretch}{0.7}

\setlength{\tabcolsep}{0.99pt}
\renewcommand{\arraystretch}{0.7}
\begin{figure*}
\centering
        \begin{tabular}{cccccccc}
            \multirow{2}{1em}{\rotatebox[origin=l]{90}{\scriptsize NASA EO}} &
            \includegraphics[width=6.3em]{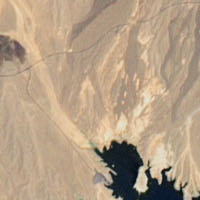}& 
            \includegraphics[width=6.3em]{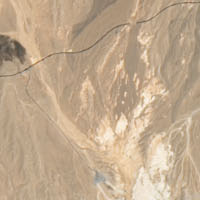}& 
            \includegraphics[width=6.3em]{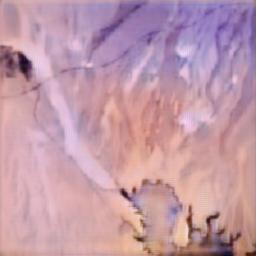}& 
            \includegraphics[width=6.3em]{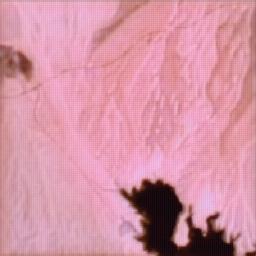}& 
            \includegraphics[width=6.3em]{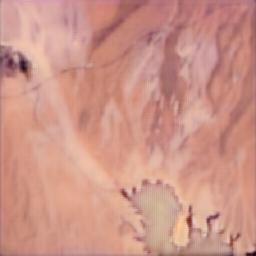}& 
            \includegraphics[width=6.3em]{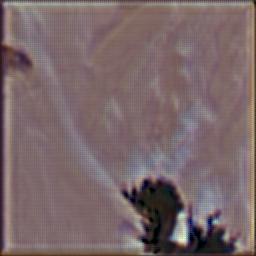} &         
            \includegraphics[width=6.3em]{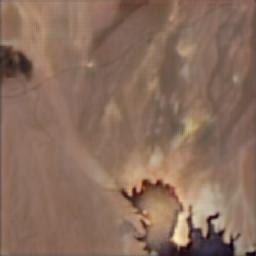} \\
            &
            \includegraphics[width=6.3em]{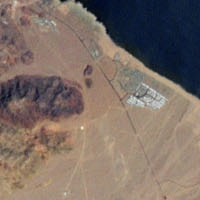}& 
            \includegraphics[width=6.3em]{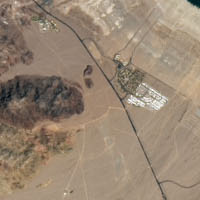}&
            \includegraphics[width=6.3em]{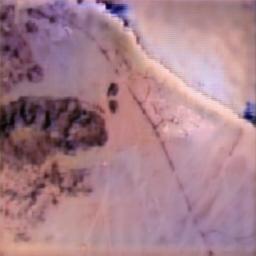}& 
            \includegraphics[width=6.3em]{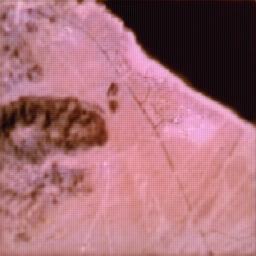}& 
            \includegraphics[width=6.3em]{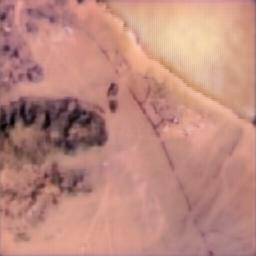} & 
            \includegraphics[width=6.3em]{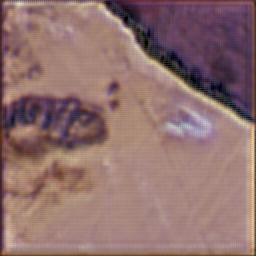} & 
            \includegraphics[width=6.3em]{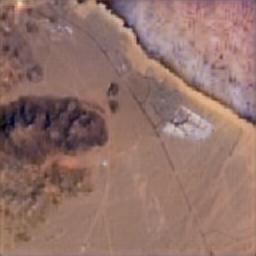}\\
             &\scriptsize Input image & \scriptsize Ground truth & \scriptsize cGANs-baseline & \scriptsize cWGAN-2 & \scriptsize cGANs-$G_{t}$ & \scriptsize cGAN-VGG &\scriptsize DichroGAN
        
        \end{tabular}
        \vspace{0.05cm}
        \caption{Sample results from ablation  on NASA EO dataset. From left-to-right: 
        Lake Mead satellite images from 2000 (input image showing high water levels) and 2022 (ground truth showing dry regions). 
        Subsequent columns show results of our ablation experiments.}
	\label{fig: Ablation Imgs}
    \vspace{-0.03cm}
    \end{figure*}

\section{Experiments and Results}

\subsection{Dataset}
We compile a dedicated dataset by extracting and organising hyperspectral observations from the PRecursore IPerSpettrale della Missione Applicativa (PRISMA) satellite. PRISMA is equipped with a visible near-infrared (VNIR) and short-wave infrared (SWIR) spectrometer, as well as a panchromatic camera. The satellite provides imagery at a spatial resolution of 30 m \cite{cogliati2021prisma}.

We use Level 2 products containing surface reflectance and 63 bands from the VNIR spectral range (400–1010 nm). RGB images are generated by selecting bands 33 (R), 45 (G), and 56 (B). Binary masks are also created to separate the land and clouds (0) from water (1).
Our dataset covers geographical regions in Australia and Mexico, comprising a total of 1,570 RGB images, along with their corresponding spectral bands (98K images) and masks.

\subsection{Implementation} \label{subsec: implementation}
DichroGAN comprises four generators and one discriminator, trained simultaneously. All generative models adopt a U-Net \cite{ronneberger2015u} architecture, while the discriminator is based on a Transformer \cite{dosovitskiy2020image}. The generators incorporate a ResNet50 \cite{he2016deep} backbone pretrained on ImageNet \cite{Deng:2009}, and include $5$ decoder blocks (256, 128, 64, 32 and 16). Generators $G_{t}$ and $G_{j}$ are initialised using  pretrained weights from \cite{gonzalez2025scene}.

Our method is implemented in PyTorch and run on an AMD EPYC 7402P (2.8GHz, 24-core) processor with 60 GB of RAM. In all the experiments, both input and output sizes are 3 × 256 × 256, except for generators $G_{s}$ and $G_{t}$, which produce outputs of 1 × 256 × 256. We train on $1,500$ images with their respective spectral bands and masks, using a batch size of $6$ and a fixed seed value of $100$. Training spans $130$ epochs, with all five networks sharing the same learning rate of $0.0002$ and momentum values of $0.5$ and $0.999$. The Adam optimiser is used for optimisation. The loss function hyperparameters are set as follows: $\gamma = 30$, $\sigma = 90$, $\iota = 100$, $\tau = 50$, and $\nu = 10$.

\subsection{Metrics}
We evaluate DichroGAN through both quantitative and qualitative analyses. First, we conduct an ablation study to examine the contribution of its components, followed by comparisons with SOTA methods. For full-reference evaluation, we use the Structural Similarity Metric (\textbf{SSIM}) \cite{wang2004image} and Peak Signal-to-Noise Ratio (\textbf{PSNR}) \cite{huynh2008scope}. For no-reference evaluation, we compare the restored satellite and underwater images using the Natural Image Quality Evaluator (\textbf{NIQE}) \cite{mittal2012making}, Underwater Image Quality Metric (\textbf{UIQM}) \cite{panetta2015human}, and \textbf{CCF} metric \cite{wang2018imaging}.

\subsection{Ablation study}

We present a comparative evaluation between the baseline and final model. The baseline model employs non-simultaneous training with two independent cGANs. In all experiments, we use a Transformer-based discriminator $D$ and apply the same loss functions for each generator. 

Recall that the dewatered and restored seafloor radiance, $\hat{y}$, is predicted by the generator $G_{j}$, while the  underwater light transmission is estimated by $G_{t}$. The surface radiance in the atmosphere, $\hat{y}_r$, is obtained by summing the diffuse and specular reflections predicted by generators $G_{d}$ and $G_{s}$. Table \ref{tab: ablation studies} summarises the ablation study and their components.

(1) \textbf{cGANs-baseline} consists of two cGANs trained separately, using binary cross-entropy as adversarial loss \cite{isola2017image}. The first cGAN (cGAN $1$) receives $x$ as input and comprises generators $G_{d}$ and $G_{s}$, their output is $\hat{y}_r$. The second cGAN (cGAN $2$) consists of one generator, $G_{j}$, which takes $\hat{y}_r$ input and generates the final prediction, $\hat{y}$. Each cGAN has a single discriminator, $D$ and $D_{1}$, respectively. 

(2) \textbf{cWGANs-2} replaces the binary cross-entropy loss with Earth Mover’s Distance by employing two conditional Wasserstein GANs (cWGANs) \cite{arjovsky2017wasserstein}, denoted as cWGAN $1$ and cWGAN $2$. All other components remain the same. 

(3) \textbf{cGANs-G$_{t}$} has cGAN $1$ unchanged. However, in cGAN $2$, we introduce a second generator, $G_{t}$, alongside $G_{j}$. With input $\hat{y}_r$, the final output remains $\hat{y}$.

(4) \textbf{cGAN-VGG} trains a single cGAN with four generators and one discriminator, $D$, in a simultaneous training. $x$ is the input for $G_{d}$, $G_{s}$, and $G_{t}$. The first two generators predict $\hat{y}_r$, while $G_{t}$ predicts $\hat{y}_t$. Generator $G_{j}$ then takes $\hat{y}_r$ as input and predicts the dewatered seafloor reflectance, $\hat{y}$. The primary objective is the optimisation of $\hat{y}$. All generators adopt a VGG \cite{simonyan2014very} architecture with a pretrained U-Net backbone \cite{gonzalez2025scene}.

\begin{table}[!t]
\fontsize{9}{10}\selectfont 
  \centering
\setlength{\tabcolsep}{3pt}
\renewcommand{\arraystretch}{1.1}
\begin{tabular}{cccccccc}
\hline
\multicolumn{8}{c}{Two cGANs separated training}\\
\hline
&\multicolumn{3}{c}{cGAN $1$} & \multicolumn{3}{c}{cGAN $2$}\\
 Model & $G_{d}$ & $G_{s}$ &  $D$ & $G_{j}$ & $G_{t}$ & $D_{1}$ & Adv. loss\\
\hline
cGANs-baseline& \checkmark & \checkmark & \checkmark &  \checkmark& & \checkmark & Cross-entropy\\
cWGANs & \checkmark & \checkmark & \checkmark &  \checkmark& & \checkmark & EMD\\
cGANs-$G_{t}$ & \checkmark & \checkmark & \checkmark &  \checkmark& \checkmark & \checkmark & Cross-entropy\\
\hline
\multicolumn{8}{c}{One cGAN simultaneous training}\\
\hline
&\multicolumn{5}{c}{cGAN}\\
Model & $G_{d}$ & $G_{s}$ &  $D$ & $G_{j}$ & $G_{t}$ & $D_{1}$ & Adv. loss\\
\hline
cGAN-VGG &  \checkmark & \checkmark & \checkmark &   \checkmark & \checkmark& & Cross-entropy\\
DichroGAN &  \checkmark & \checkmark & \checkmark &   \checkmark & \checkmark& & Cross-entropy\\
\bottomrule
\end{tabular}
\vspace{0.05cm}
\caption{Ablation study. The first three models consist of two cGANs trained separately, whereas the last two models consist of one cGAN with four generators and one discriminator.}
\label{tab: ablation studies}
\vspace{-0.3cm}
\end{table}

\begin{table}[!b]
\fontsize{9}{10}\selectfont 
\centering
\setlength{\tabcolsep}{3pt}
\renewcommand{\arraystretch}{1}
\begin{tabular}{cc c   }
\hline
Ablation & $SSIM\uparrow$ & $ PSNR\uparrow$\\
\hline
cGANs-baseline& 0.593 & 17.75 \\
cWGANs & 0.524 & 17.96\\
cGANs-$G_{t}$ & 0.582 & 17.78\\
cGAN-VGG & 0.489 & 16.34\\
DichroGAN & \textbf{0.672} & \textbf{18.01}\\
\bottomrule
\end{tabular}
\vspace{0.02cm}
\caption{Quantitative evaluation of the ablation study.}
\label{tab: ablation results}
\end{table}

\setlength{\tabcolsep}{3pt}
\renewcommand{\arraystretch}{0.8}
\begin{table}
\fontsize{9}{10}\selectfont 
  \centering
\renewcommand{\arraystretch}{1.1}
\begin{tabular}{cccc}
\hline

 Method & $SSIM\uparrow$ & $PSNR\uparrow$ & $NIQE\downarrow$ \\
\hline
UDCP \cite{drews2016underwater}& 0.511 & 11.90 & \textbf{4.656}
\\
CWR \cite{han2022underwater}& 0.536 & 13.66 & \textbf{4.857}  \\
NU$^2$Net \cite{guo2023underwater}& 0.535 & 12.87 & 5.439\\
Phaseformer \cite{Khan_2025_WACV}& \textbf{0.554} & \textbf{13.80} & 6.056 \\
Ours &\textbf{0.560} & \textbf{14.39} & 5.643 \\

\bottomrule
\end{tabular}
\vspace{0.05cm}
\caption{Quantitative evaluation on NASA EO. 
}
\label{tab:results Lake Mead}
\vspace{-0.3cm}
\end{table}

\setlength{\tabcolsep}{3pt}
\renewcommand{\arraystretch}{0.8}
\begin{table}[!t]
\small
  \centering
\renewcommand{\arraystretch}{1}
\begin{tabular}{c c c c}
\hline
 Method&$CCF\uparrow$ & $UIQM\uparrow$ & $NIQE\downarrow$ \\
\hline

Input image & 10.79 & 1.556  & 5.813 \\
UDCP \cite{drews2016underwater}& \textbf{24.72} & 1.395  & 6.624  \\
CWR \cite{han2022underwater}&  16.91 & \textbf{2.807}  & \textbf{4.924}  \\
NU$^2$Net \cite{guo2023underwater} & 18.50 & \textbf{2.878}  & 5.746 \\
Phaseformer \cite{Khan_2025_WACV} &10.81 & 2.435  & 5.759 \\
Ours &  \textbf{18.84} & 2.342  & \textbf{5.422} \\

\bottomrule
\end{tabular}
\vspace{0.05cm}
\caption{Quantitative evaluation on PRISMA and NASA EO.
}
\label{tab:results satellite datasets and methods}
\vspace{-0.3cm}
\end{table}

\setlength{\tabcolsep}{3pt}
\renewcommand{\arraystretch}{0.8}
\begin{table}[!b]
\fontsize{9}{10}\selectfont 
  \centering
\renewcommand{\arraystretch}{1}
\begin{tabular}{c c c cc}
\hline
 Method &$CCF\uparrow$ & $UIQM\uparrow$ & $NIQE\downarrow$ \\
\hline
Input image (HICDR) & 14.50 & 3.091& 9.598 \\ 
UDCP \cite{drews2016underwater}& \textbf{33.78} &  2.861&  10.31   \\
CWR \cite{han2022underwater}& \textbf{43.25} & 3.327 & \textbf{5.918}  \\
NU$^2$Net \cite{guo2023underwater}& 29.09 & \textbf{3.591} & 8.047   \\
Phaseformer \cite{Khan_2025_WACV}& 22.57 & \textbf{3.637} &9.488  \\
Ours &  25.93  & 3.351  &  \textbf{5.889} \\
\hline
Input image (UIEB) & 27.50 & 2.760 & 4.394 \\
UDCP \cite{drews2016underwater}& \textbf{49.71} & 2.317& 5.474  \\
CWR \cite{han2022underwater}& 20.33 & 2.975& \textbf{3.990}  \\
NU$^2$Net \cite{guo2023underwater}& 26.21 & \textbf{3.278}& 4.451  \\
Phaseformer \cite{Khan_2025_WACV}& 16.42 & \textbf{3.327}& 5.181  \\
Ours &  \textbf{26.41} & 3.005& \textbf{3.987}  \\

\bottomrule
\end{tabular}
\vspace{0.05cm}
\caption{Quantitative evaluation on  HICRD \cite{han2022underwater} and UIEB \cite{li2019underwater}.
}
\label{tab:results underwater datasets and methods}
\vspace{-0.3cm}
\end{table}

Table \ref{tab: ablation results} shows the quantitative results of our ablation using satellite imagery from NASA Earth Observatory (EO) program. We select images of Lake Mead in the Southwestern United States, focusing on the years 2000 and 2022 to illustrate changes in water levels and exposed terrain. 

DichroGAN achieves the highest SSIM and PSNR. Fig. \ref{fig: Ablation Imgs} provides a qualitative comparison, showing that the binary cross-entropy adversarial loss and U-Net networks yields better performance. Among the two non-simultaneously trained cGANs, adding the fourth generator, $G_{t}$, responsible for light transmission estimation, improves water column removal, though it slightly alters the scene's overall colour.
In contrast, DichroGAN closely identifies and removes water regions to recover the in-air colours of the seafloor.

\subsection{Comparison with SOTA}
To the best of our knowledge, no existing  methods specifically remove water regions and restore the in-air RGB colours of seafloor from satellite imagery. Therefore, to ensure a fair comparison, we apply our method on both satellite and underwater images to validate our experiments. 

For comparative analysis, we use SOTA for underwater image enhancement and restoration, including UDCP \cite{drews2016underwater}, CWR \cite{han2022underwater}, NU$^2$Net \cite{guo2023underwater}, and Phaseformer \cite{Khan_2025_WACV}, and a combination of satellite and underwater datasets, including  a testing set from PRISMA \cite{cogliati2021prisma}, a dataset from NASA's EO program, the Underwater Image Enhancement Benchmark (UIEB) \cite{li2019underwater}, and the Heron Island Coral Reef Dataset (HICRD) \cite{han2022underwater}. In total, we use 654 images for comparison.

\begin{figure*}[!t]
        \centering
        \begin{tabular}{ccccccccc}
       \rotatebox[origin=l]{90}{\scriptsize LAKE MEADE}
            \includegraphics[width=6.3em]{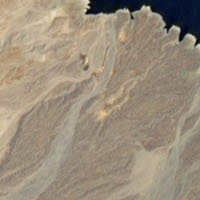}&
            \includegraphics[width=6.3em]{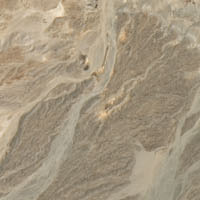}& 
            \includegraphics[width=6.3em]{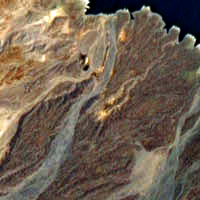}& 
            \includegraphics[width=6.3em]{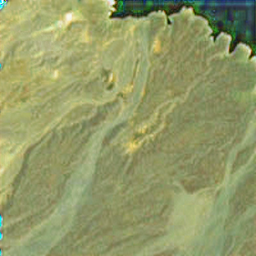}& 
            \includegraphics[width=6.3em]{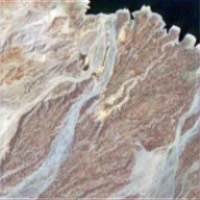} & 
            \includegraphics[width=6.3em]{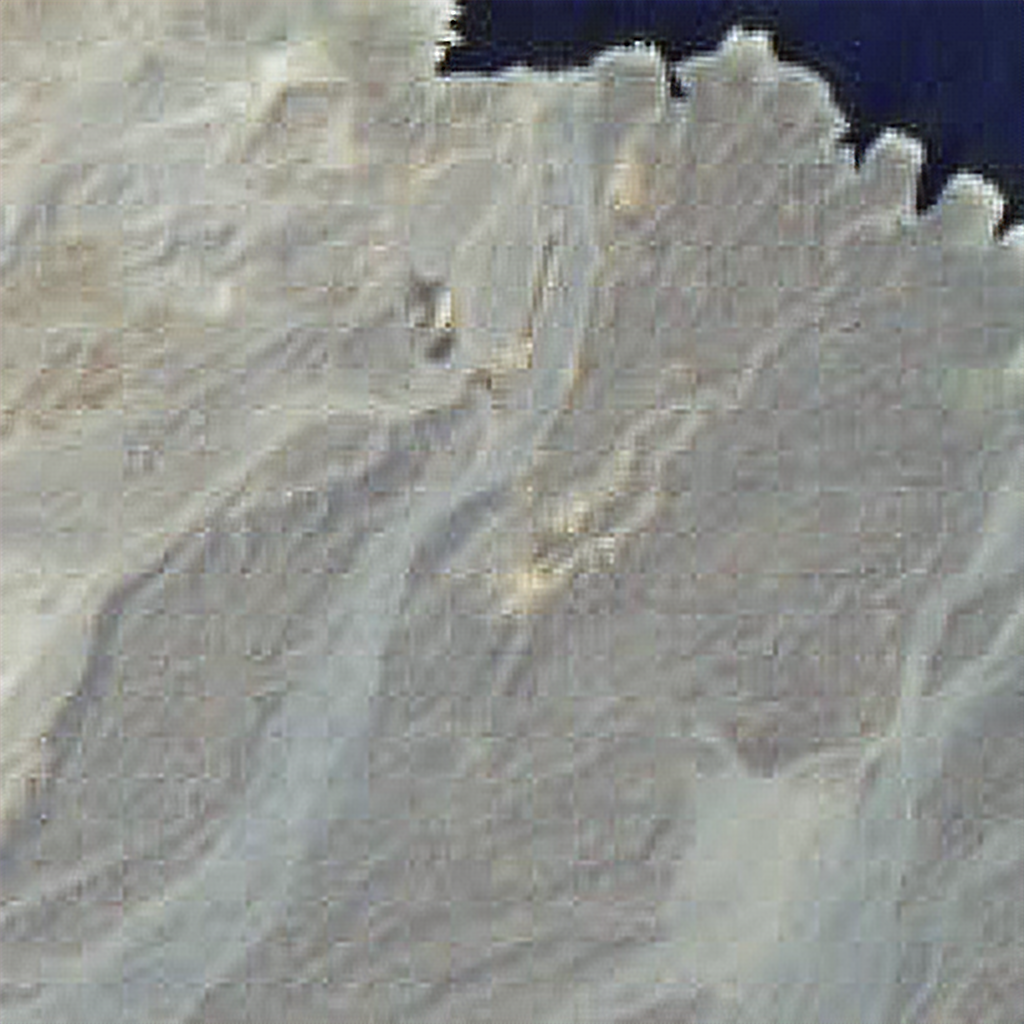} & 
            \includegraphics[width=6.3em]{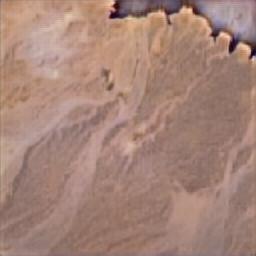}\\
            
             \scriptsize Input image &\scriptsize Ground truth & \scriptsize UDCP \cite{drews2016underwater} & \scriptsize CWR \cite{han2022underwater} & \scriptsize NU$^2$Net \cite{guo2023underwater} & \scriptsize Phaseformer \cite{Khan_2025_WACV} & \scriptsize Ours
        
        \end{tabular}
        \vspace{0.05cm}
        \caption{Sample results on Lake Meade images from NASA EO. 
        }
	\label{fig: Lake Mead images}
    \vspace{-0.1cm}
    \end{figure*}

Table \ref{tab:results Lake Mead} shows the quantitative results for full-reference metrics on Lake Mead images (NASA EO). Our method achieves the highest SSIM and PSNR, where higher values indicate better performance. UDCP method performs best for image quality according to the NIQE metric, where lower values are better. However, in the qualitative comparison, our method effectively removes the water region as shown in Figs. \ref{fig: Ablation Imgs} and \ref{fig: Lake Mead images}.

In Table \ref{tab:results satellite datasets and methods}, we present the results for non-reference metrics on the NASA EO and PRISMA datasets. For the CCF and UIQM metrics, higher numbers indicate better performance, while for the NIQE metric, a value closest to zero represents the best performance. Our method achieves the second-best performance for the CCF and NIQE metrics, while UDCP and NU$^2$Net  obtain the best scores, respectively. Table \ref{tab:results underwater datasets and methods} shows the results on the underwater datasets HICRD and UIEB . For these datasets, our method achieves the best NIQE score, while UDCP, CWR, and Phaseformer obtain the best scores for CCF and UIQM metrics.

\begin{figure}[!t]
        \centering
        \begin{tabular}{ccccccc}
            \multirow{3}{1em}{\rotatebox[origin=l]{90}{\scriptsize PRISMA}}&
            \includegraphics[width=3.5em]{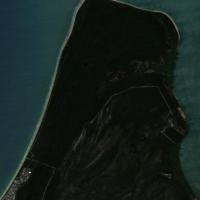}& 
            \includegraphics[width=3.5em]{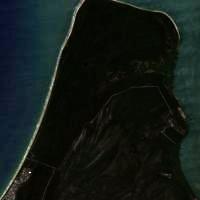}& 
            \includegraphics[width=3.5em]{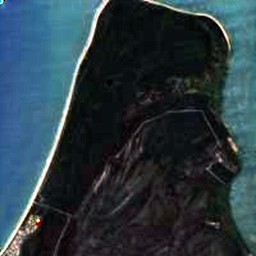}& 
            \includegraphics[width=3.5em]{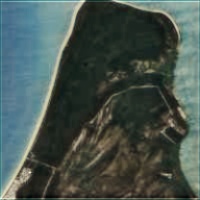}& 
            \includegraphics[width=3.5em]{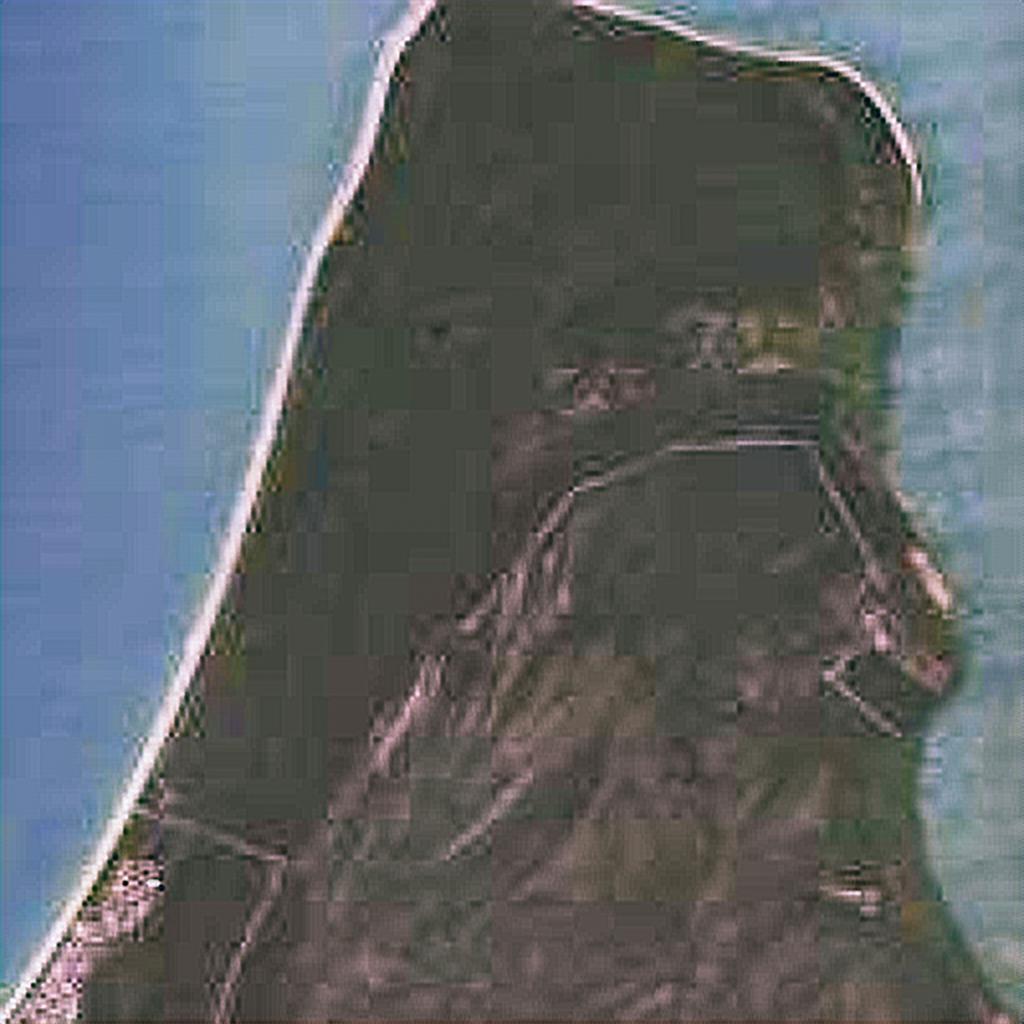} & 
            
            \includegraphics[width=3.5em]{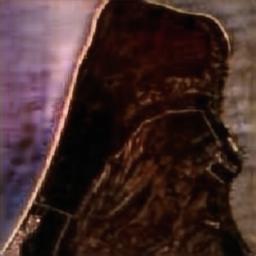} \\
            &
            \includegraphics[width=3.5em]{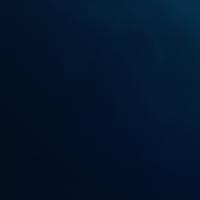}& 
            \includegraphics[width=3.5em]{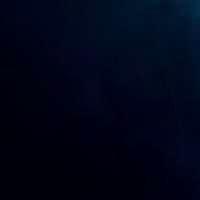}& 
            \includegraphics[width=3.5em]{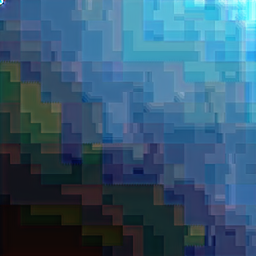}& 
            \includegraphics[width=3.5em]{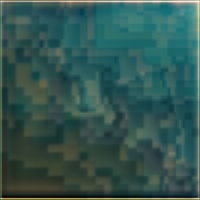} & 
            \includegraphics[width=3.5em]{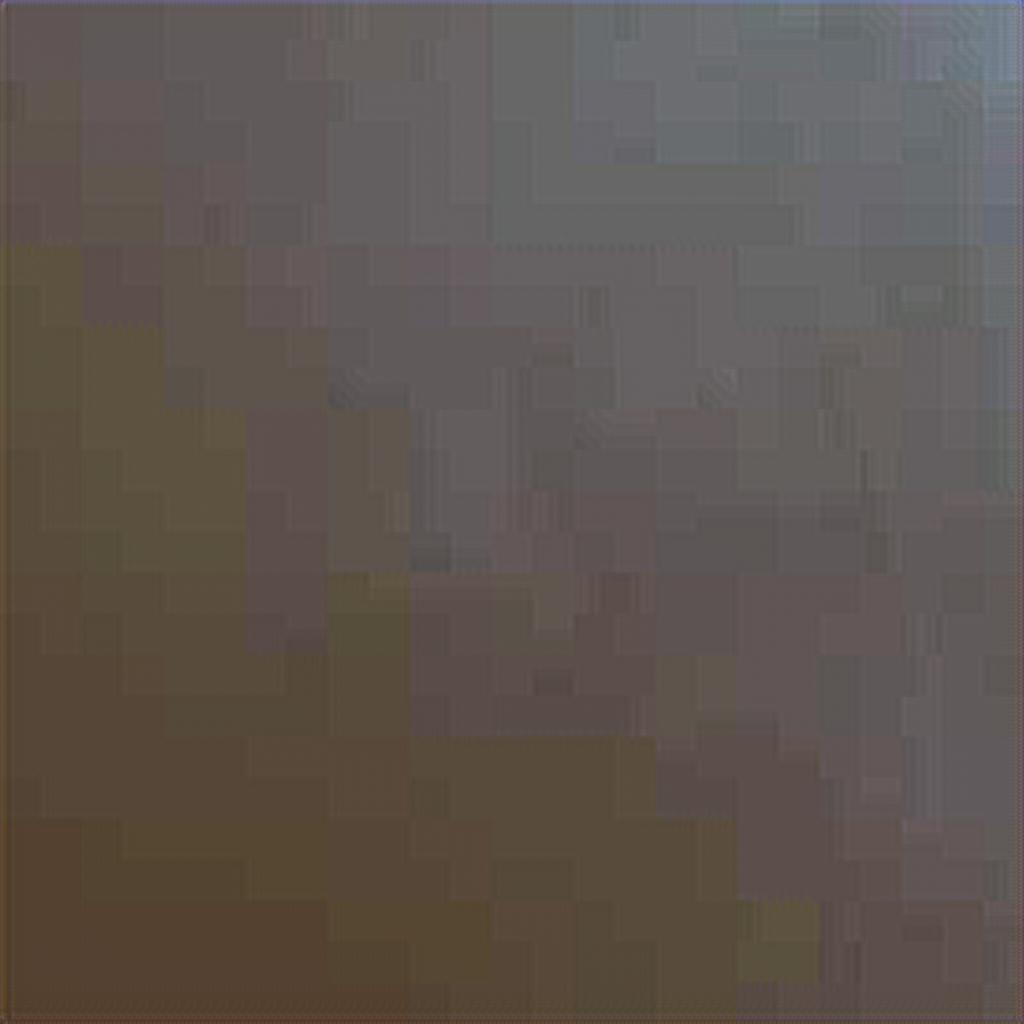} & 
            \includegraphics[width=3.5em]{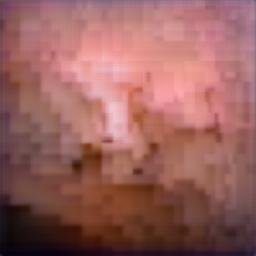}\\
            \multirow{3}{1em}{\rotatebox[origin=l]{90}{\scriptsize NASA EO}}&
            \includegraphics[width=3.5em]{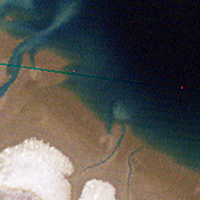}& 
            \includegraphics[width=3.5em]{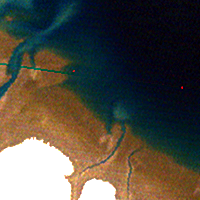}& 
            \includegraphics[width=3.5em]{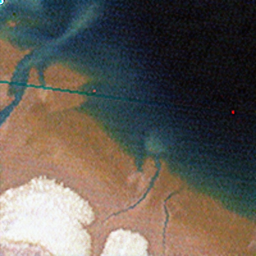}& 
            \includegraphics[width=3.5em]{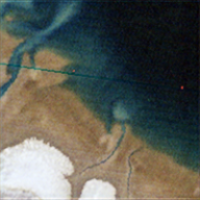}& 
            \includegraphics[width=3.5em]{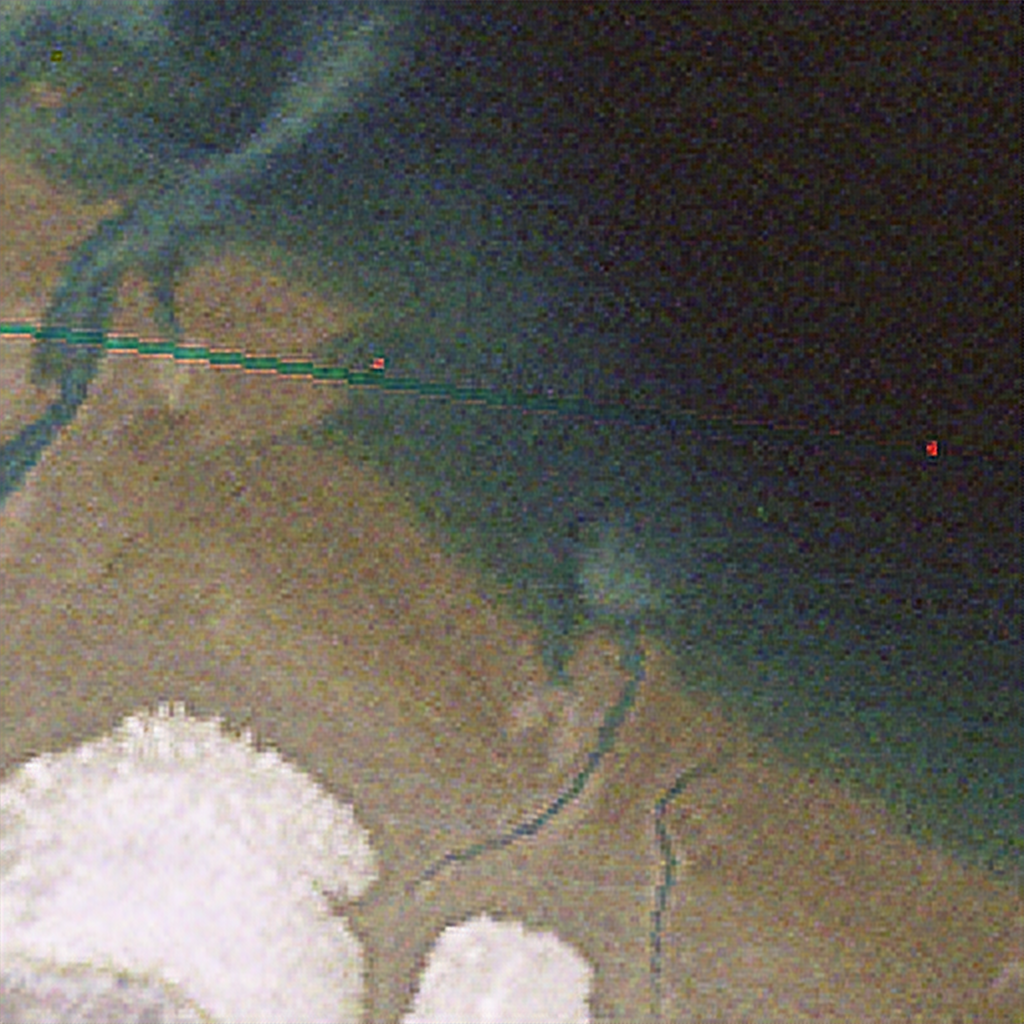} & 
            \includegraphics[width=3.5em]{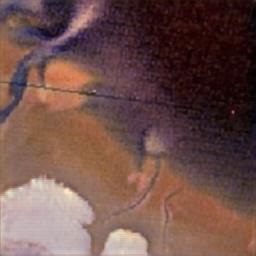} \\
            &
            \includegraphics[width=3.5em]{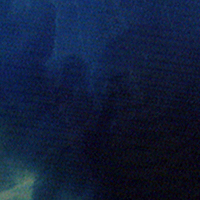}& 
            \includegraphics[width=3.5em]{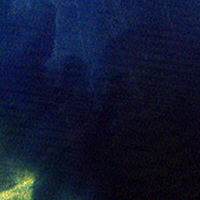}& 
            \includegraphics[width=3.5em]{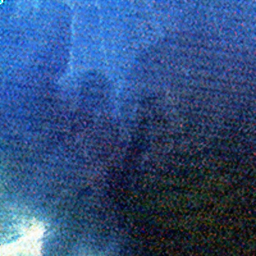}& 
            \includegraphics[width=3.5em]{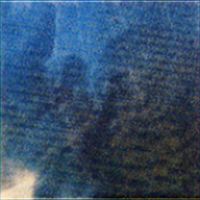} & 
            \includegraphics[width=3.5em]{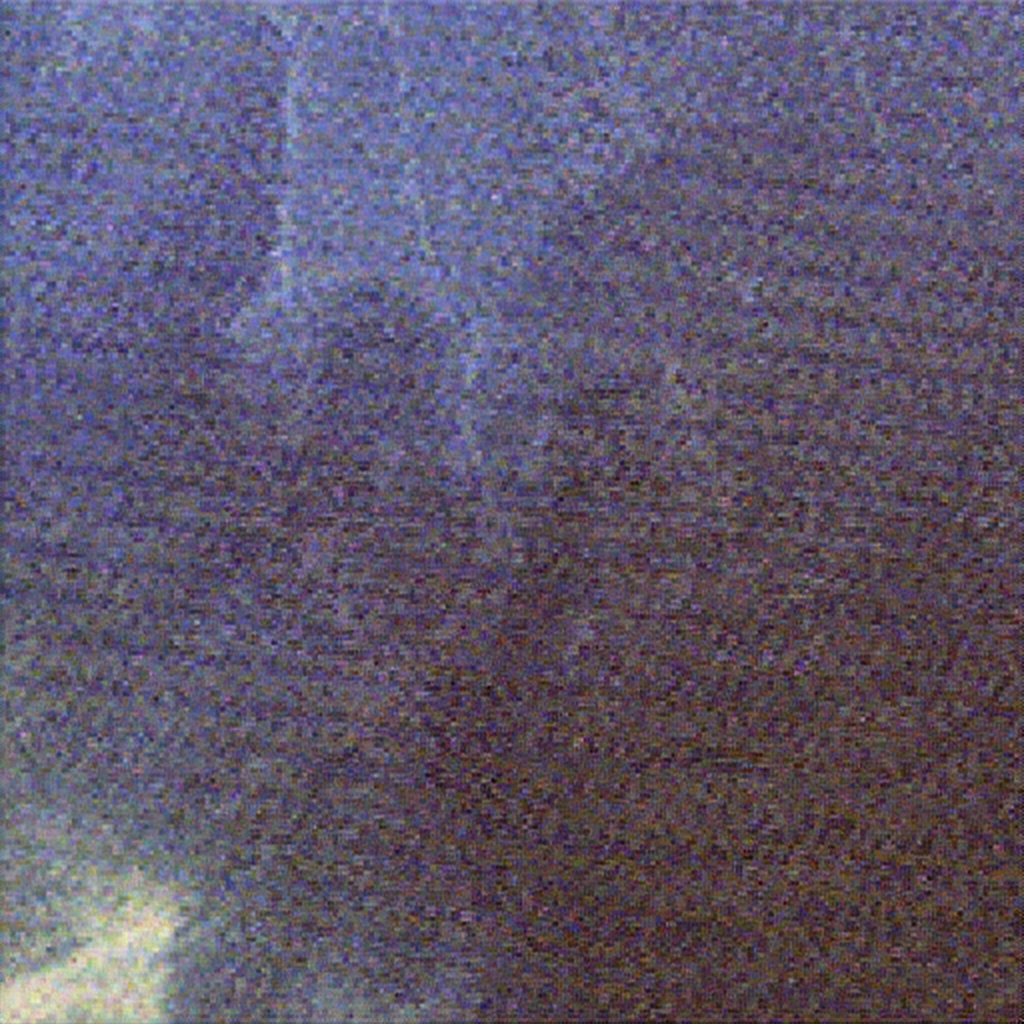} & 
            \includegraphics[width=3.5em]{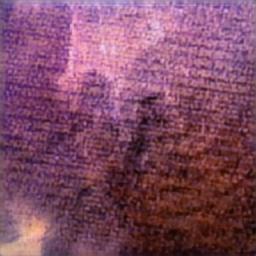}\\
            
             &\scriptsize Input image & \scriptsize UDCP 
             & \scriptsize CWR 
             & \scriptsize NU$^2$Net 
             & \scriptsize Phaseformer 
             & \scriptsize Ours
        
        \end{tabular}
        \vspace{0.05cm}
        \caption{Sample results on PRISMA and NASA EO. 
        }
	\label{fig: PRIMSA images}
    \vspace{-0.1cm}
    \end{figure}

\begin{figure}[!b]
        \centering
        \begin{tabular}{ccccccc}
            \multirow{3}{0.9em}{\rotatebox[origin=l]{90}{\scriptsize HICRD}}&
            \includegraphics[width=3.5em]{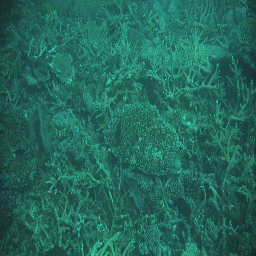}& 
            \includegraphics[width=3.5em]{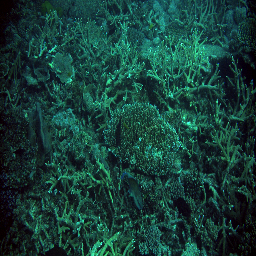}& 
            \includegraphics[width=3.5em]{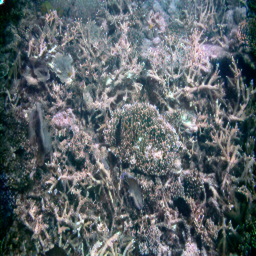}& 
            \includegraphics[width=3.5em]{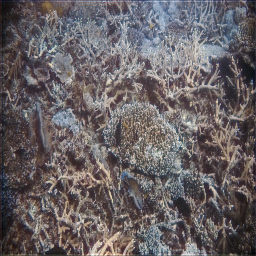}& 
            \includegraphics[width=3.5em]{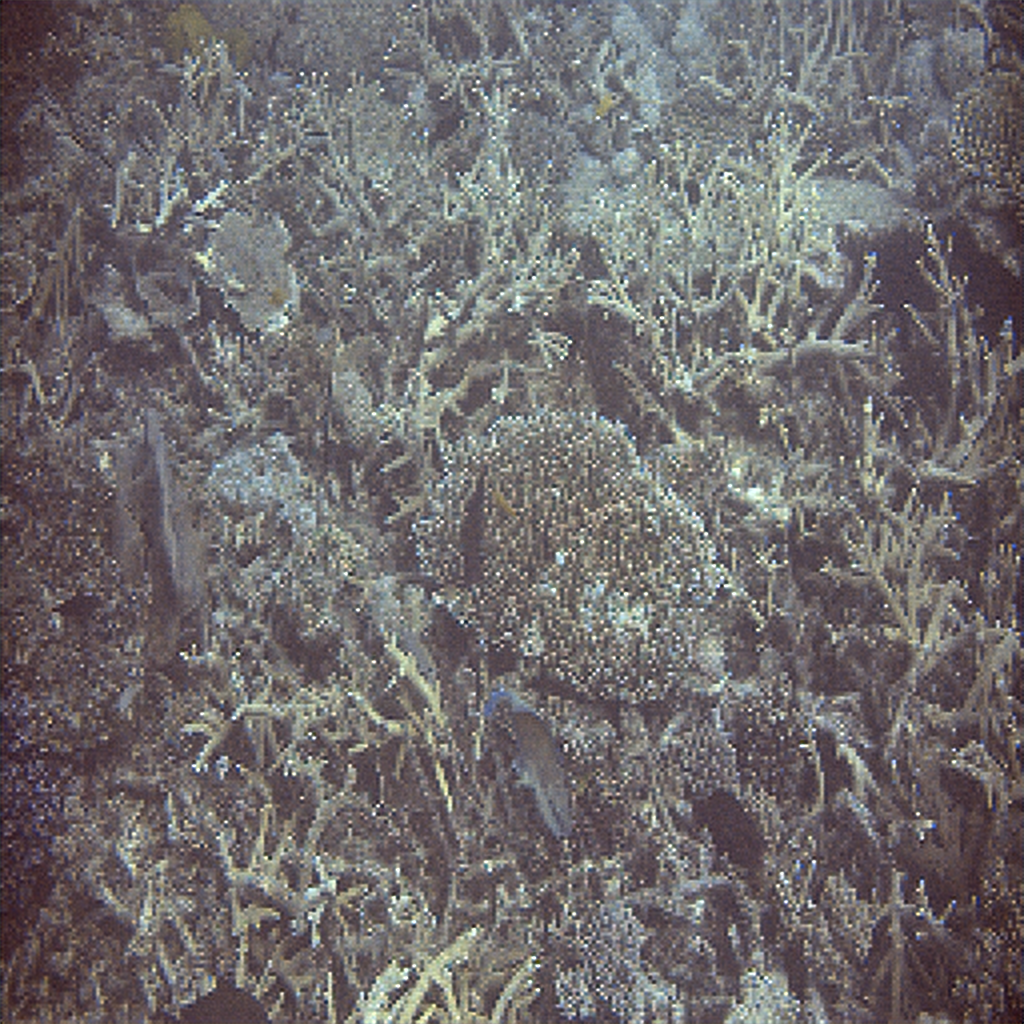} & 
            
            \includegraphics[width=3.5em]{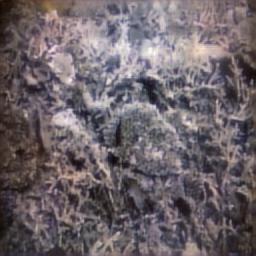} \\
            &
            \includegraphics[width=3.5em]{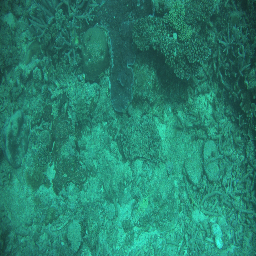}& 
            \includegraphics[width=3.5em]{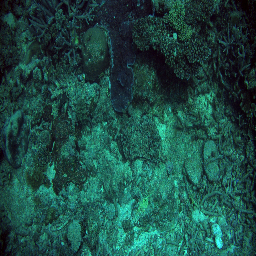}& 
            \includegraphics[width=3.5em]{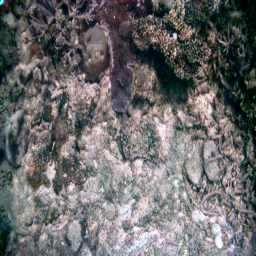}& 
            \includegraphics[width=3.5em]{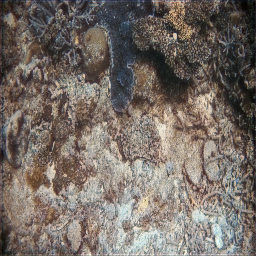} & 
            \includegraphics[width=3.5em]{imgs/NY_test165.png} & 
            \includegraphics[width=3.5em]{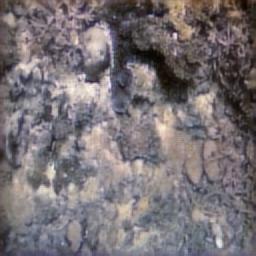}\\
            \multirow{3}{0.9em}{\rotatebox[origin=l]{90}{\scriptsize UIEB}}&
            \includegraphics[width=3.5em]{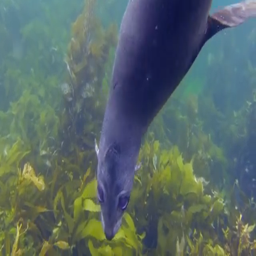}& 
            \includegraphics[width=3.5em]{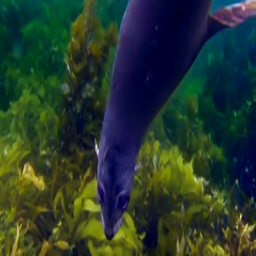}& 
            \includegraphics[width=3.5em]{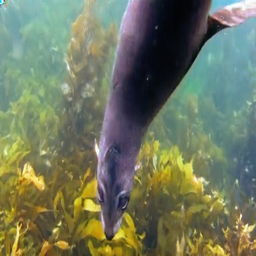}& 
            \includegraphics[width=3.5em]{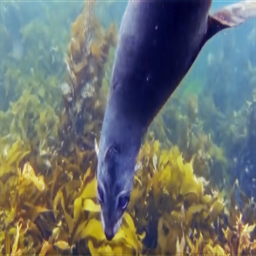}& 
            \includegraphics[width=3.5em]{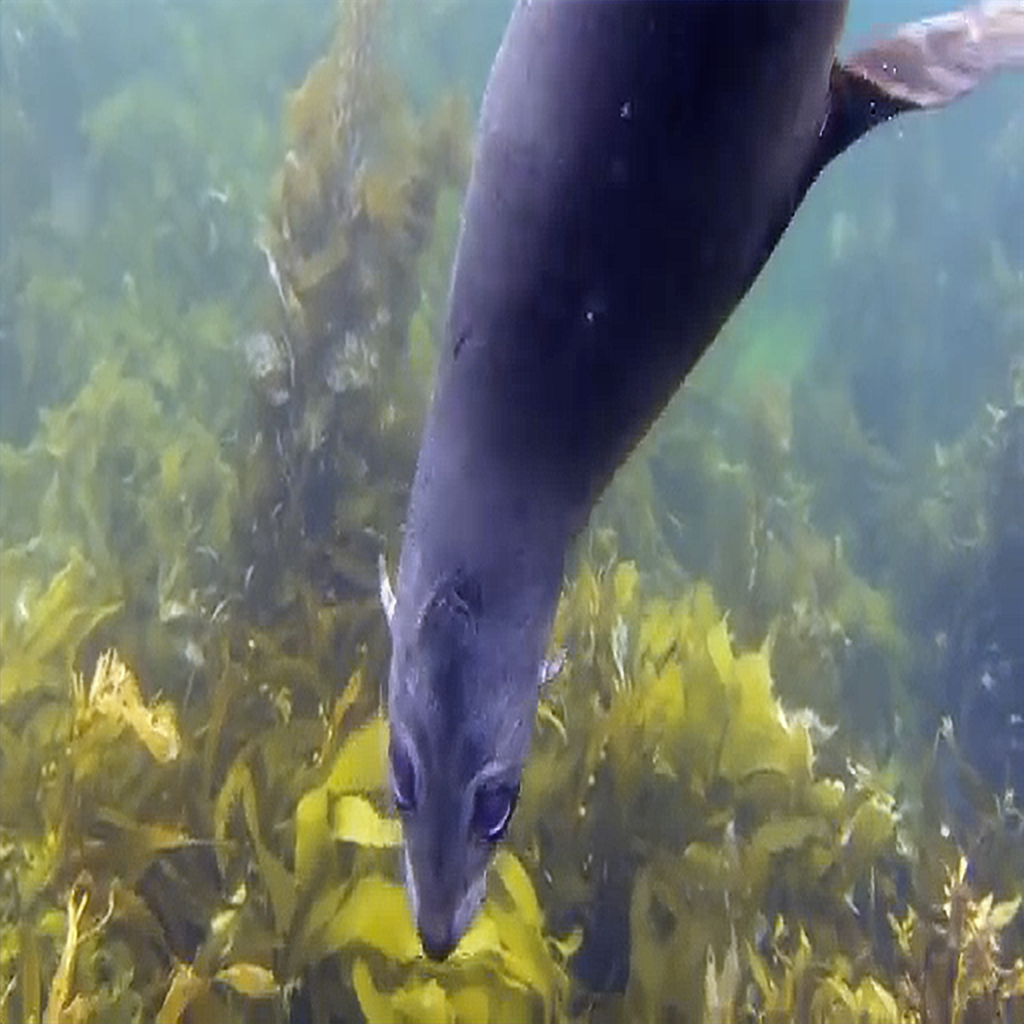} & 
            \includegraphics[width=3.5em]{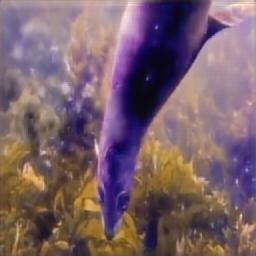} \\
            &
            \includegraphics[width=3.5em]{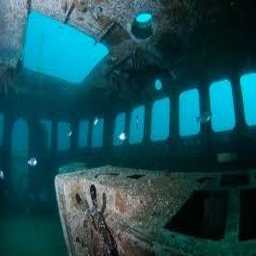}& 
            \includegraphics[width=3.5em]{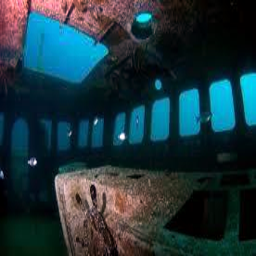}& 
            \includegraphics[width=3.5em]{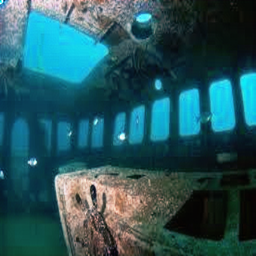}& 
            \includegraphics[width=3.5em]{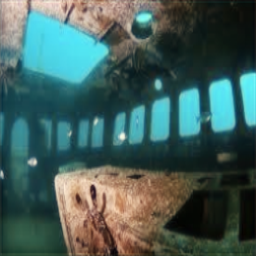} & 
            \includegraphics[width=3.5em]{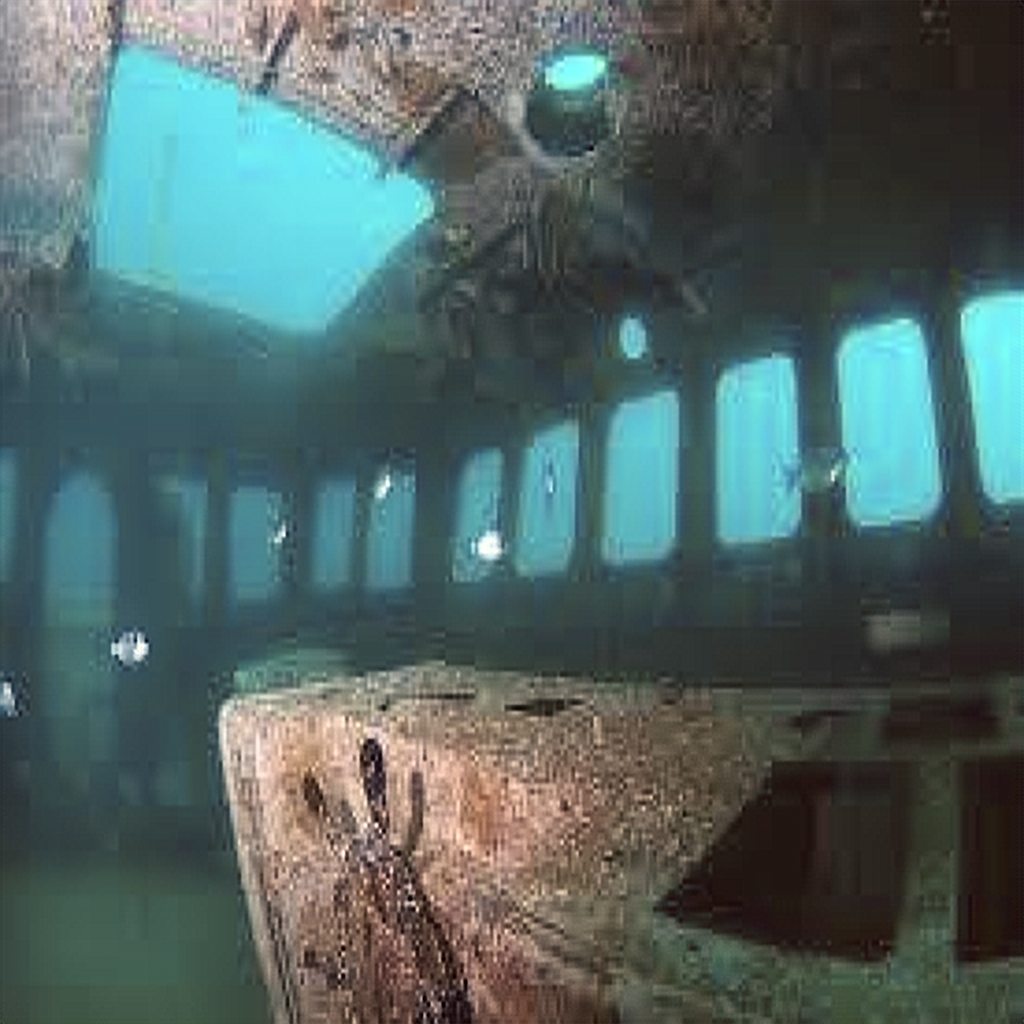} & 
            \includegraphics[width=3.5em]{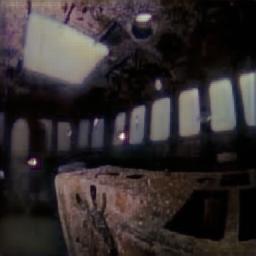}\\
            
             &\scriptsize Input image & \scriptsize UDCP 
             & \scriptsize CWR 
             & \scriptsize Nu$^2$Net 
             & \scriptsize Phaseformer 
             & \scriptsize Ours
        
        \end{tabular}
        \vspace{0.05cm}
        \caption{Sample results on HICRD \cite{han2022underwater} and UIEB \cite{li2019underwater}. 
        }
	\label{fig: Hicrd and uieb images}
    \vspace{-0.1cm}
    \end{figure}

In Fig. \ref{fig: Lake Mead images}, we compare the ground truth images from Lake Meade with the generated images. The results align with the quantitative analysis, demonstrating that our method successfully removes the water column while preserving  the image structure. The other methods fail  to accurately identify the water regions. 
Results on the PRISMA and NASA EO datasets are shown in Fig. \ref{fig: PRIMSA images}. Our method attempts to remove the water and recover the seafloor's in-air colour, providing a clearer representation of the underlying terrain. Fig. \ref{fig: Hicrd and uieb images} shows examples on HICRD and UIEB  datasets.While UDCP achieves higher scores on the underwater metrics, it does not remove the colour cast. 

\setlength{\tabcolsep}{0.99pt}
\renewcommand{\arraystretch}{0.6}

\setlength{\tabcolsep}{0.7pt}
\renewcommand{\arraystretch}{0.5}
    
\setlength{\tabcolsep}{0.7pt}
\renewcommand{\arraystretch}{0.5}

\subsection{Discussion}

Visual inspection confirms only DichroGAN removes the colour cast. However, as a physics-based method focused on accurate colour recovery, it underperforms on standard underwater metrics that tend to favour over-enhanced images \cite{li2019underwater}. Despite achieving reasonable consistency, it also exhibits a degree of blurriness, which is a common GAN limitation attributed to difficulties in learning high-frequency details. Future work might focus on improving performance by increasing the dataset size and employing multi-term loss optimisation to enhance texture and high-resolution mapping.

\section{Conclusion}
In this work, we introduce DichroGAN, a novel cGAN designed to recover the in-air colours of seafloor from satellite imagery. By employing a two-step simultaneous training process, DichroGAN effectively estimates diffuse and specular reflections, computes atmospheric scene radiance, and models underwater light transmission to mitigate the effects of light absorption and scattering. Experimental results demonstrate the promising performance of our method across a range of satellite and underwater datasets. 

{
\footnotesize
\bibliographystyle{ieee}
\bibliography{./egbib}
}

\end{document}